\documentclass[runningheads]{llncs}
\usepackage[T1]{fontenc}
\usepackage{graphicx}
\usepackage{multirow}
\usepackage{tabularx}
\usepackage{array}
\newcolumntype{C}[1]{>{\centering\arraybackslash}p{#1}}
\usepackage{amsmath}
\usepackage{amsfonts}
\usepackage[dvipsnames]{xcolor}
\usepackage[section]{placeins}

\begin{document}
\title{Home Run: Finding Your Way Home by Imagining Trajectories}
%
%

\author{Daria de Tinguy\inst{1} \and
Pietro Mazzaglia\inst{1} \and
Tim Verbelen\inst{1} \and 
Bart Dhoedt\inst{1}}%

\institute{
IDLab, Department of Information Technology \\
Ghent University - imec \\ 
Technologiepark-Zwijnaarde 126, B-9052 Ghent, Belgium \\
\email{{firstname.lastname}@ugent.be}
}

\authorrunning{D. de Tinguy \& al.}

%
%
\maketitle              
\begin{abstract}
When studying unconstrained behavior and allowing mice to leave their cage to navigate a complex labyrinth, the mice exhibit foraging behavior in the labyrinth searching for rewards, returning to their home cage now and then, e.g. to drink. Surprisingly, when executing such a ``home run'', the mice do not follow the exact reverse path, in fact, the entry path and home path have very little overlap. Recent work proposed a hierarchical active inference model for navigation, where the low level model makes inferences about hidden states and poses that explain sensory inputs, whereas the high level model makes inferences about moving between locations, effectively building a map of the environment. However, using this ``map'' for planning, only allows the agent to find trajectories that it previously explored, far from the observed mice's behaviour. In this paper, we explore ways of incorporating before-unvisited paths in the planning algorithm, by using the low level generative model to imagine potential, yet undiscovered paths. We demonstrate a proof of concept in a grid-world environment, showing how an agent can accurately predict a new, shorter path in the map leading to its starting point, using a generative model learnt from pixel-based observations.

\end{abstract}

\section{Introduction}

Humans rely on an internal representation of the environment to navigate, i.e. they do not require precise geometric coordinates or complete mappings of the environment; a few landmarks along the way and approximate directions are enough to find our way back home \cite{cognitive_map_graph}. This reflects the concept of a ``cognitive map'' as introduced by Tolman~\cite{Tolman1948CognitiveMI}, and matches the discovery of specific place cells firing in the rodent hippocampus depending on the animal position \cite{1stratslam} and our representation of space \cite{cognitive_map_graph}. 

Recently, \c{C}atal et al.~\cite{hierarchical_AIF} showed how such mapping, localisation and path integration can naturally emerge from a hierarchical active inference (AIF) scheme and are also compatible with the functions of the hippocampus and entorhinal cortex~\cite{Safron2021}. This was implemented on a real robot to effectively build a map of its environment, which could then be used to plan its way using previously visited locations~\cite{latentslam}.

However, while investigating the exploratory behaviour of mice in a maze, where mice were left free to leave their home to run and explore, a peculiar observation was made. When the mice decided to return to their home location, instead of re-tracing their way back, the mice were seen taking fully new, shorter, paths directly returning them home~\cite{mice_exp}.

On the contrary, when given the objective to reach a home location, the hierarchical active inference model, as proposed by~\cite{hierarchical_AIF,latentslam}, can only navigate between known nodes of the map, unable to extrapolate possible new paths without first exploring the environment. To address this issue, we propose to expand the high level map representation using the expected free energy of previously unexplored transitions, by exploiting the learned low-level environment model. In other worlds, we enlarge the projection capabilities of architecture~\cite{latentslam} to unexplored paths.

In the remainder of this paper we will first review the hierarchical AIF model \cite{hierarchical_AIF}, then explain how we address planning with previously unvisited paths by imagining novel trajectories within the model. As a proof of concept, we demonstrate the mechanism on a Minigrid environment with a four-rooms setup. We conclude by discussing our results, the current limitations and what is left to improve upon the current results.


\section{Navigation as hierarchical active inference}

The active inference framework relies upon the notion that intelligent agents have an internal (generative) model optimising beliefs (i.e. probability distributions over states), explaining the causes of external observations. By minimising the surprise or prediction error, i.e, free energy (FE), agents can both update their model as well as infer actions that yield preferred outcomes \cite{FristonActInfLearning,nav_aif}.

\begin{figure}[t!]
    \centering
    \includegraphics[width=2.5in]{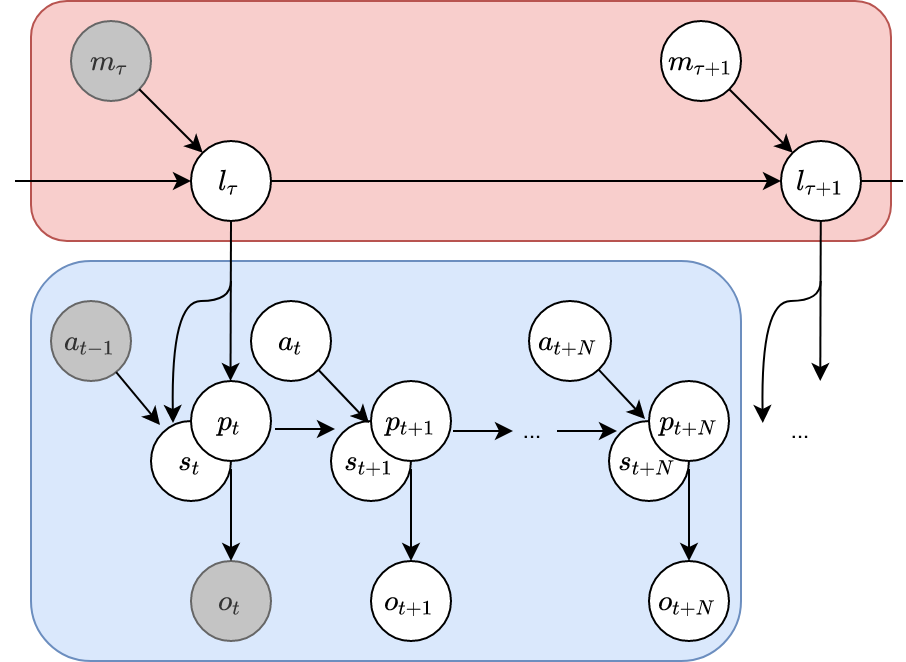}
    \caption{Navigation as a hierarchical generative model for active inference \cite{hierarchical_AIF}. At the lower level, highlighted in blue, the model entertains beliefs about hidden states $s_{t}$ and $p_{t}$, representing hidden causes of the observation and the pose at the current timestep $t$ respectively. The hidden states give rise to observations $o_t$, whereas actions $a_t$ impact future states. At the higher level, highlighted in red, the agent reasons about locations $l$. The next location $l_{
    \tau+1}$ is determined by executing a move $m_{
    \tau}$ . Note that the higher level operates on a coarser timescale. Grey shaded nodes are considered observed.} 
    \label{fig:model}
\vspace{-3mm}
\end{figure}

In the context of navigation, \c{C}atal et al.~\cite{hierarchical_AIF} introduced a hierarchical active inference model, where the agent reasons about the environment on two different levels. On the low level, the agent integrates perception and pose, whereas on the high level the agent builds a more coarse grained, topological map. This is depicted in Figure~\ref{fig:model}.

The low level, depicted in blue, comprises a sequence of low-level action commands $a_t$ and sensor observations $o_t$, which are generated by hidden state variables $s_t$ and $p_t$. Here $s_t$ encodes learnable features that give rise to sensory outcomes, whereas $p_t$ encodes the agent's pose in terms of its position and orientation. The low level transition model $p(s_{t+1} | s_{t}, p_{t}, a_{t})$ and likelihood model $p(o_t | s_t)$ are jointly learnt from data using deep neural networks~\cite{generative_model}, whereas the pose transition model $p(p_{t+1} | s_{t}, p_{t}, a_{t})$ is instantiated using a continuous attractor network similar to \cite{ratslam1}.

At the high level, in red in the Figure, the agent reasons over more coarse grained sequences of locations $l_\tau$, where it can execute a move $m_\tau$ that gives rise to a novel location $l_{\tau+1}$. In practice, this boils down to representing the environment as a graph-based map, where locations $l_\tau$ are represented by nodes in the graph, whereas potential moves $m_\tau$ are links between those nodes. Note that a single time step at the higher level, i.e. going from $\tau$ to $\tau + 1$, can comprise multiple time steps on the lower level.  This enables the agent to first `think' far ahead in the future on the higher level.

To generate motion, the agent minimizes expected free energy (EFE) under this hierarchical generative model. To reach a preferred outcome, the agent first plans a sequence of moves that are expected to bring the agent to a location rendering the preferred outcome highly plausible, after which it can infer the action sequence that brings the agent closer to the first location in that sequence. For a more elaborate description of the generative model, the (expected) free energy minimisation and implementation, we refer to \cite{hierarchical_AIF}. 

\section{Imagining unseen trajectories}

As discussed in \cite{hierarchical_AIF}, minimising expected free energy under such a hierarchical model induces desired behaviour for navigation. In the absence of a preferred outcome, an epistemic term in the EFE will prevail, encouraging the agent to explore actions that yield information on novel (hidden) states, effectively expanding the map while doing so. In the presence of a preferred state, the agent will exploit the map representation to plan the shortest (known) route towards the objective. However, crucially, the planning is restricted to previously visited locations in the map. This is not consistent with the behaviour observed in mice~\cite{mice_exp}, as these, apparently, can exploit new paths even when engaging in a goal-directed run towards their home.

In order to address this issue, we hypothesize that the agent not only considers previously visited links and locations in the map during planning, but also imagines potential novel links. A potential link from a start location $l_A$ to a destination location $l_B$ is hence scored by the minimum EFE over all plans $\pi$ (i.e. a sequence of actions) generating such a trajectory under the (low level) generative model, i.e.:

\begin{equation} 
\begin{split}
	G(l_A, l_B)
	    &= \min_\pi \sum_{k=1}^{H} \underbrace{D_{KL}\big[Q(s_{t+k}, p_{t+k} | \pi)Q(s_t | l_A) \Vert Q(s_{t+H}, p_{t+H} | l_{B}) \big]}_\text{probability reaching $l_B$ from $l_A$} \\ 
	    & + \underbrace{\mathbb{E}_{Q(s_{t+k})}\big[H(P(o_{t+k} | s_{t+k}))\big]}_\text{observation ambiguity}.
\end{split}
\label{eq:G_low}
\vspace{-3mm}
\end{equation}

The first term is a KL divergence between the expected states to visit starting at location $l_A$ and executing plan $\pi$, and the state distribution expected at location $l_B$. The second term penalizes paths that are expected to yield ambiguous observations.

We can now use $G(l_A,l_B)$ to weigh each move between two close locations (the number of path grows exponentially the further the objective is), even through ways not explored before, and plan for the optimal trajectory towards a goal destination. In the next section, we work out a practical example using a grid-world environment.


\section{Experiments}

\subsection{MiniGrid setup}


\begin{figure}[!h]
\begin{tabular}{lll} 
    {A)}\begin{minipage}{0.5\textwidth} \includegraphics[width=1\textwidth]{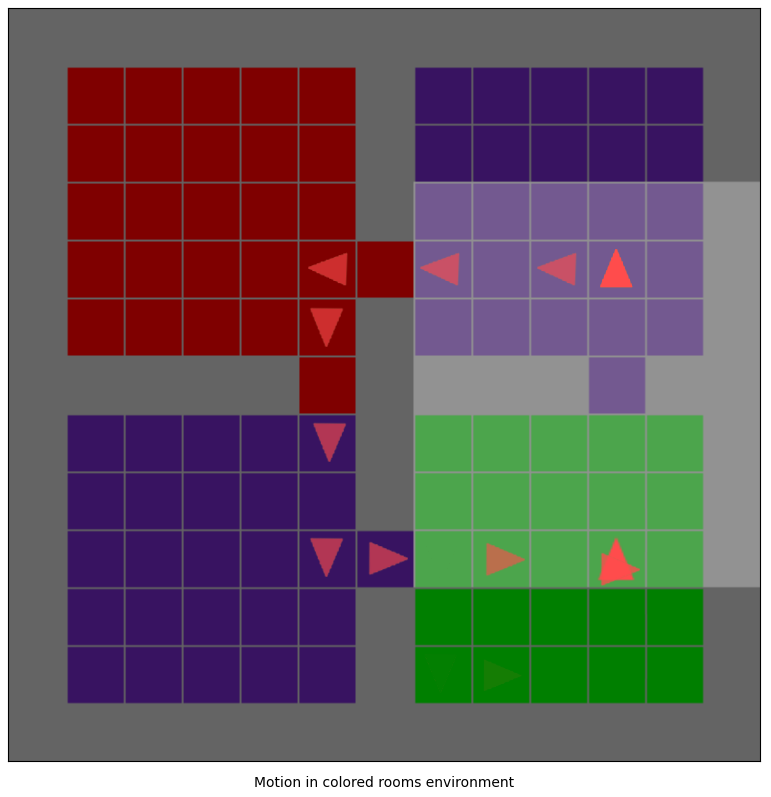}\label{fig:a} \end{minipage} &  \begin{minipage}{0.5\textwidth}
    {B)}\includegraphics[width=0.7\textwidth]{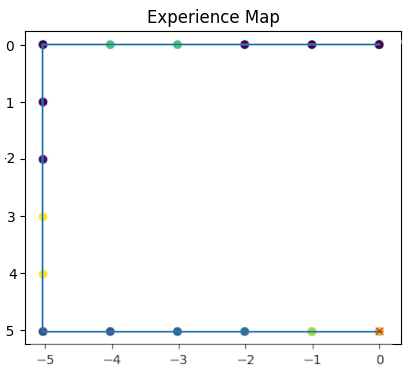}\label{fig:b} \\
    {C)}\includegraphics[width=0.7\textwidth]{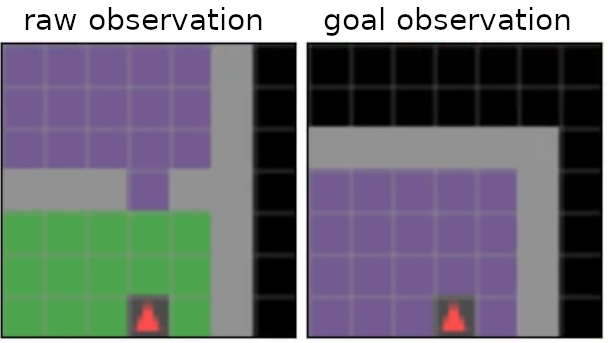}\label{fig:c}  
    
    \end{minipage}
\end{tabular}
\caption{MiniGrid test maze and associated figures, A) An example of the maze with a reachable goal (door open allowing shortcut) and the agent path toward a home-run's starting point, the transparent grey box correspond to the agent's field of view at the starting position. B) The topological map of the path executed in A as generated by the high level of our generative model, C) The currently observed RGB image as reconstructed by the agent's model at the end of path and the view at the desired goal position.}
\label{fig:setup}
\vspace{-3mm}
\end{figure} 

The experiments were realised in a MiniGrid environment \cite{gym_minigrid} of 2$\times$2 up to 5$\times$5 rooms, of sizes going from 4 to 7 tiles and having a random floor color chosen among 6 options : red, green, blue, purple, yellow and grey. Rooms are connected by a single open tile, randomly spawned in the wall. The agent has 3 possible actions at each time step: move one tile forward, turn 90 degrees left or turn 90 degrees right.
It can't see through walls and can only venture into an open grid space. Note that the wall blocking vision is not really realistic and the agent can see the whole room if there is an open door in its field of view, thus even if part of the room should be masked by a wall (eg. Fig \ref{fig:setup}C raw observation).
It can see ahead and around in a window of $7\times7$ tiles, including its own occupied tile. The observation the agent receives is a pixel rendering in RGB of shape $3 \times 56 \times 56$. 


\subsection{Model training and map building}

Our hierarchical generative model was set up in similar fashion as~\cite{hierarchical_AIF}. To train the lower level of the generative model, which consists of deep neural networks, we let an agent randomly forage the MiniGrid environments, and train those end to end by minimising the free energy on those sequences. Additional model details and training parameters can be found in Appendix~\ref{appendix:model}.

The high level map is also built using the same procedure as~\cite{hierarchical_AIF}. However, since we are dealing with a grid-world, distinct places in the grid typically yield distinct location nodes in the map, unless these are near and actually yield identical observations. Also, we found that predicting the effect of turning left or right was harder for neural networks to predict, yielding a higher surprise signal. However, despite these limitations, we can still demonstrate the main contribution of this paper.

\subsection{Home run}

Inspired by the mice navigation in \cite{mice_exp}, we test the following setup in which the agent first explores a maze, and at some point is provided with a preference of returning to the start location. Figure~\ref{fig:setup} shows an example of a test environment and associated trajectories realised by the agent. At the final location, the agent is instructed to go back home, provided by the goal observation in Fig.~\ref{fig:setup}C. Fig.~\ref{fig:setup}B illustrates the map generated by the hierarchical model.

\begin{figure}[t!]
    \centering
    \includegraphics[width=0.9\textwidth]{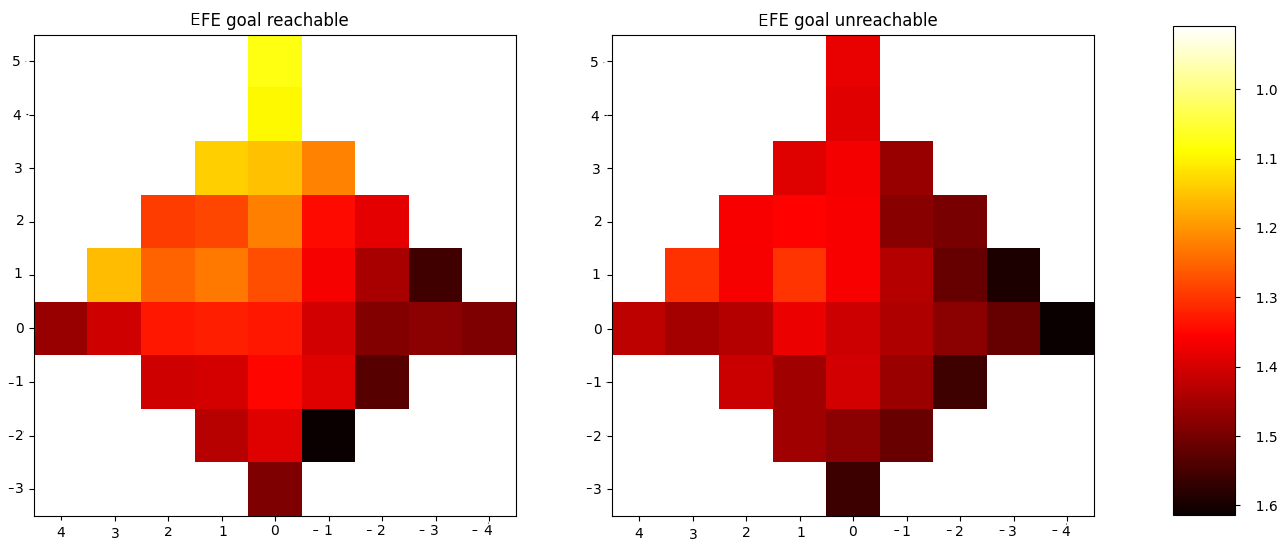}
    \caption{Lowest expected free energy of each end position after 5 steps. The right figure shows the agent at position (0,0) facing the goal at position (0,5), as represented in Figure~\ref{fig:setup}A i). In the left figure, the door is open, therefore the goal is reachable, on the right figure the door is closed, the goal cannot be reached in 5 steps.}
    \label{fig:EFE}
    \vspace{-4mm}
\end{figure}

First, we test whether the agent is able to infer whether it can reach the starting node in the experience map from the current location. We do so by imagining all possible plans $\pi$, and evaluating the expected free energy of each plan over an average of $N=3$ samples from the model. Figure~\ref{fig:EFE} shows the EFE for all reachable locations in a 5 steps planning horizon. It is clear that in case the door is open, the agent expects the lowest free energy when moving forward through the door, expecting to reach the start node in the map. In case the path is obstructed (the door as in \ref{fig:setup}A, allowing a shortcut, is closed), it can still imagine going forward 5 steps, but this will result in the agent getting stuck against the wall, which it correctly imagines and reflects on the EFE.

\begin{figure}[b!]
    \vspace{-10mm}
    \centering
    \includegraphics[width=0.8\textwidth]{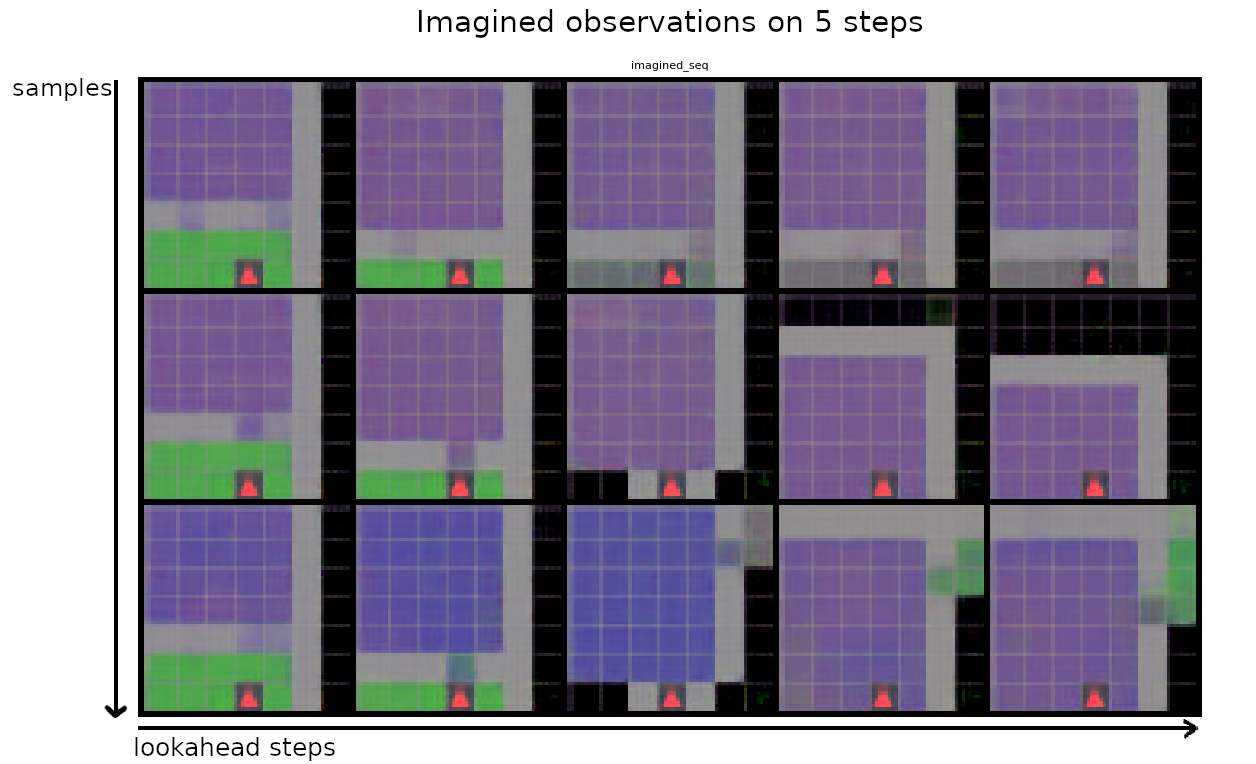}
    \caption{Three imagined trajectories of a 5-steps projection moving forward. The trained model is not perfectly predicting the future, only the middle sequence predicts the correct dynamics.}
    \label{fig:im5steps}
    \vspace{-5mm}
\end{figure}

However, the prior model learnt by the agent is far from perfect. When inspecting various imagined rollouts of the model, as shown in Figure~\ref{fig:im5steps}, we see that the model has trouble encoding and remembering the exact position of the door, i.e. predicting the agent getting stuck (top) or incorrect room colours and size (bottom). While not problematic in our limited proof of concept, also due to the fact that the EFE is averaged over multiple samples, this shows that the effectiveness of the agent will be largely dependent on the accuracy of the model. 

To test the behaviour in a more general setting, we set multiple home-run scenarios, where the agent's end position is $d=5,6,7,9$ steps away from the start location. For each $d$, we sample at least 20 runs over 4 novel $2\times2$ rooms environment, with different room sizes and colours, similar to the train set, in which 10 have an open door between the start and goal, and 10 have not. We count the average number of steps required by the agent to get back home, and compare against two baseline approaches. First is the Greedy algorithm, inspired by~\cite{vector_nav}, in which the agent greedily navigates in the direction of the goal location, and follows obstacles in the known path direction when bumping into one. Second is a TraceBack approach, which retraces all its steps back home, similar to Ariadne's thread. Our approach uses the EFE with a planning horizon of $d$ to decide whether or not the home node is reachable based on a fixed threshold, and falls back to planning in the hierarchical model, which boils down to a TraceBack strategy.

\begin{table}[t!]
\begin{tabular}{p{0.7cm}|C{1.7cm}C{1.7cm}C{1.7cm}|C{1.7cm}C{1.7cm}C{1.7cm}}
  & \multicolumn{3}{c}{\textbf{open}}  & \multicolumn{3}{c}{\textbf{closed}} \\ 
$d$ & Greedy & TraceBack & Ours & Greedy  & TraceBack & Ours \\ \hline
5 & 5      & 25        & 6.5    & 29.5      & 25        & 25   \\
6 & 6      & 31        & 6    & 41      & 31        & 31   \\
7 & 7      & 27        & 11.5 & 31.5      & 27        & 27   \\
9 & 9      & 36        & 23.7   & 46      & 36        & 36  
\end{tabular}
\caption{Home run strategies and the resulting number of steps, for different distances $d$ to home, and open versus closed scenarios. For small $d$ our model correctly imagines the outcome. For $d=9$ the agent infers an open door about 27\% of the time.}
\label{tab:homeruns}
\vspace{-4mm}
\end{table}

In case of small $d$ ($\leq$6), our approach successfully identifies whether the goal is reachable or not, even when the agent is not facing it, which results in a similar performance for a Greedy approach in the `open' case, and a reverting to TraceBack in the `closed' case. There is been only one exception in our test-bench at 5steps range issued by a reconstruction error on all samples (the occurrence probability is 0.04\%  as having a sample wrongly estimating the door position at 5steps is 33\%). For $d=7$ our model misses some of the shortcut opportunities, as the model's imagination becomes more prone to errors for longer planning horizons. For $d=9$, the rooms are larger and the wall separating the two rooms is actually not visible to the agent. In this regime, we found the agent imagines about 27\% of the time that it will be open, and takes the gamble to move towards the wall, immediately returning on its path if the wall is obstructed.


\section{Discussion}

Our experiments show that using the EFE of imagined paths can yield more optimal, goal-directed behaviour. Indeed, our agent is able to imagine and exploit shortcuts when planning its way home. However, our current experimental setup is still preliminary and we plan to further expand upon this concept. For instance we currently arbitrarily set the point at which the agent decide to home-run. In a real experiment, the mice likely decide to go home due to some internal stimulus, e.g., when they get thirsty and head back home where water is available. We could further develop the experimental setup to incorporate such features and do a more extensive evaluation.

One challenge of using the Minigrid environment as an experimental setup \cite{gym_minigrid} is the use of top view visual observations. Using a pixel-wise error for learning the low-level perception model can be problematic, as for example the pixel-wise error between a closed versus an open tile in the wall is small in absolute value, and hence it's difficult to learn for the model, as illustrated in Figure~\ref{fig:im5steps}. A potential approach to mitigate this is to use a contrastive objective instead, as proposed by~\cite{contrastive}.

Another important limitation of the current model is that it depends on the effective planning horizon of the lowest level model to imagine shortcuts. Especially in the Minigrid environment, imagining the next observation for a 90 degree turn is challenging, as it requires a form of memory of the room layout to correctly predict the novel observation. This severely limits the planning horizon of our current models. A potential direction of future work in this regard is to learn a better location, state and pose mapping. For instance, instead of simply associating locations with a certain state and pose, conditioning the transition model on a learnt location descriptor might allow the agent to learn and encode the shape of a complete room in a location node.

Other approaches have been proposed to address the navigation towards a goal by the shortest way possible in a biologically plausible way. For instance, Erdem et al. \cite{spiking_nav} reproduced the pose and place-cell principle of the rat's hippocampus with spiking neural networks and
use a dense reward signal to drive goal-directed behaviour, with more reward given the closer the agent gets to the goal. Hence, the path with the highest reward is sought, and trajectories on which obstacles are detected are discarded.
In Vegard et al. \cite{vector_nav}, the process is also bio-inspired, based on the combination of grid cell-based vector and topological navigation. The objective is now explicitly represented as a target position in space, which is reached by vector navigation mechanisms with local obstacle avoidance mediated by border cells and place cells. 
Both alternatives also adopt topological maps and path integration in order to reach their objective. However, both exhibit more greedy and reactive behaviour, whereas our model is able to exploit the lower level perception model to already predict potential obstacles upfront, before bumping into those.   




\section{Conclusion}


In this paper we have proposed how a hierarchical active inference model can be used to improve planning by predicting novel, previously unvisited paths. We demonstrated a proof of concept using a generative model learnt from pixel based observations in a grid-world environment.

As future work we envision a more extensive evaluation, comparing shallow versus deep hierarchical generative models in navigation performance. Moreover, we aim to address several of the difficulties of our current perception model, i.e. the limitations of pixel-wise prediction errors, the limited planning horizon, and a more expressive representation for locations in the high level model. Ultimately, our goal is to deploy this on a real-world robot, autonomously exploring, planning and navigating in its environment.

\section*{Acknowledgment}
This research received funding from the Flemish Government under the “Onder-
zoeksprogramma Artificiële Intelligentie (AI) Vlaanderen” programme.

\bibliographystyle{IEEEtran} 

\bibliography{main} 
\appendix
\section{Model details and training}
\label{appendix:model}

In this appendix, we provide some additional details on the training data, model parameters, training procedure and building the hierarchical map.

\subsection{Training data}

To optimize the neural network models a dataset composed of sequences of action-observation pairs was collected by human demonstrations of interaction with the environment. The agent was made to move around from rooms to room, circle around and turn randomly. About 12000 steps were recorded in 39 randomly created environments having different room size, number of rooms, open door emplacements and floor colors, as well as the agent having a random starting pose and orientation.  2/3 of the data were used for training and 1/3 for validation. Then a fully novel environment was used for testing. 

\subsection{Model parameters}

The low level perception model is based on the architecture of~\cite{generative_model}, and is composed of 3 neural networks that we call: prior, posterior and likelihood.

\textbf{The prior} neural network consists in a LSTM layer followed with a variational layer giving out a distribution (i.e mean and std).

\textbf{The posterior}  model first consists of a convolutional network to compress sensor data. This data is then concatenated with the hot encoded action and the previous state, all of that is then processed by a fully connected neural network coupled with a variational layer to obtain a distribution.

\textbf{The likelihood} model performs the inverse of the convolutional part of the posterior, generating an image out of a given state sample. 

The detailed parameters are listed in Table~\ref{table:params}.

\begin{table}[!h]
\centering
\scalebox{0.7}{
\begin{tabular}{llcc}
                                               & \textbf{Layer} & \multicolumn{1}{l}{\textbf{Neurons/Filters}} & \multicolumn{1}{l}{\textbf{Stride}} \\ \hline
\multirow{3}{*}{Prior}                         & Concatenation  &                                              &                                     \\
                                               & LSTM           & 200                                          &                                     \\
                                               & Linear         & 2*30                                         &                                     \\ \hline
\multicolumn{1}{c}{\multirow{8}{*}{Posterior}} & Convolutional  & 16                                           & 2                                   \\
\multicolumn{1}{c}{}                           & Convolutional  & 32                                           & 2                                   \\
\multicolumn{1}{c}{}                           & Convolutional  & 64                                           & 2                                   \\
\multicolumn{1}{c}{}                           & Convolutional  & 128                                          & 2                                   \\
\multicolumn{1}{c}{}                           & Convolutional  & 256                                          & 2                                   \\
\multicolumn{1}{c}{}                           & Concatenation  &                                              &                                     \\
\multicolumn{1}{c}{}                           & Linear         & 200                                          &                                     \\
\multicolumn{1}{c}{}                           & Linear         & 2*30                                         &                                     \\ \hline
\multirow{12}{*}{Likelihood}                   & Linear         & 200                                          &                                     \\
                                               & Linear         & 256*2*2                                      &                                     \\
                                               & Upsample       &                                              &                                    \\
                                               & Convolutional  & 128                                          & 1                                   \\
                                               & Upsample       &                                              &                                     \\
                                               & Convolutional  & 64                                           & 1                                   \\
                                               & Upsample       &                                              &                                     \\
                                               & Convolutional  & 32                                           & 1                                   \\
                                               & Upsample       &                                              &                                     \\
                                               & Convolutional  & 16                                           & 1                                   \\
                                               & Upsample       &                                              &                                     \\
                                               & Convolutional  & 3                                            & 1                                   \\ \hline
\end{tabular}
}
\caption{Models parameters}
\label{table:params}
\end{table}

\subsection{Training the model}

The low level perception pipeline was trained end to end on time sequences of 10 steps using stochastic gradient descent with the minimization of the free energy loss function \cite{generative_model}:

\begin{flalign*}
FE & = \sum_t D_{KL}[Q(s_t | s_{t-1}, a_{t-1}, o_t) || P(s_t | s_{t-1}, a_{t-1})] - \mathbb{E}_{Q(s_t)}[\log P(o_t| s_t)]
\end{flalign*}

The loss consists of a negative log likelihood part penalizing the error on reconstruction, and a KL-divergence between the posterior and the prior distributions on a training sequence. We trained the model for 300 epochs using the ADAM optimizer~\cite{adam_optimiser} with a learning rate of 1$\cdot$10–4.




\subsection{Building the map}

The high level model is implemented as a topological graph representation, linking pose and hidden state representation to a location in the map. Here we reuse the LatentSLAM implementation~\cite{latentslam} consisting of pose cells, local view cells and an experience map.

\textbf{The pose cells} are implemented as a Continuous Attractor Network (CAN), representing the local position x, y and heading $\theta$ of the agent. Pose cells represent a finite area, therefore the firing fields of a single grid cell correspond to several periodic spatial locations. 

\textbf{The local view cells} are organised as a list of cell, each cell containing a hidden state representing an observation, the pose cell excited position, and the map's experience node linked to this view. After each motion, the encountered scene is compared to all previous cells observation by calculating the cosine distance between hidden state features. If the distance is smaller than a given threshold, then the cell corresponding to this view is activated, else a new cell is created.

\textbf{The experience map} contains the experience of the topological map. It gives an estimate of the agent global pose in the environment and link the pose cell position with the local view cell active at this moment. If those elements do not match with any existing node of the map, a new one is created and linked to the previous experience, else a close loop is operated and the existing experiences are linked together.

\end{document}